# Adaptive Motion Planning with Artificial Potential Fields Using a Prior Path

Javad Amiryan

Department of Computer Engineering
Sharif University of Technology
Tehran, Iran
amiryanj@ce.sharif.edu

Mansour Jamzad

Department of Computer Engineering
Sharif University of Technology
Tehran, Iran
jamzad@sharif.edu

*Abstract*— Motion planning in an autonomous agent is responsible for providing smooth, safe and efficient navigation. Many solutions for dealing this problem have been offered, one of which is, Artificial Potential Fields (APF). APF is a simple and computationally low cost method which keeps the robot away from the obstacles in environment. However, this approach suffers from trapping in local minima of potential function and then fails to produce motion plans. Furthermore, Oscillation in presence of obstacles or in narrow passages is another disadvantage of the method which makes it unqualified for many planning problems. In this paper we aim to resolve these deficiencies by a novel approach which employs a *prior path* between origin and goal configuration of the robot. Therefore, the planner guarantees to lead the robot to goal area while the inherent advantages of potential fields remain. For path planning stage, we intend to use randomized sampling methods such as Rapidly-exploring Random Trees (RRT) or its derivatives, however, any path planning approach can be utilized. We have also designed an optimization procedure for evolving the motion plans towards optimal solution. Then genetic algorithm is applied to find smoother, safer and shorter plans. In our experiments, we apply a simulated vehicle in Webots simulator to test and evaluate the motion planner. Our experiments showed our method to enjoy improving the performance and speed in comparison to basic approaches.

## I. INTRODUCTION

The accelerating use of robots in different aspects of modern life underlines the prominence of research in this field. Mobile robots are being employed in cleaning houses, hospital services military applications and so on. Besides giant automobile manufacturers are trying to replace human drivers by virtual intelligent drivers. Autonomous vehicles are expected to prove their competency in order for their presence on streets. Natural human-like motions which are safe and optimum are among such expectations. Central to a virtual driver is its motion planner which plays a significant role by providing a sequence of feasible motions which navigate the automobile to the destination without collision with any obstacle. Since the advent of researches on this problem, numerous solutions have been offered, however a recent challenge is to produce an optimal motion plan for online decision making: a plan which minimizes one or multiple cost functions of the user including path length, passengers' comfort and path safety.

Combinatorial planning methods are among classifications introduced in [1] that use all information in the environment and are complete. On the other hand are planning methods based on randomized sampling which incrementally search for finding a plan.

In complete motion planning often a method is applied for discretization of configuration space. Cell decomposition method is one of the simplest ways in which a connectivity graph is constructed by dividing each dimension of the space into multiple parts [2]. As an obtained path in such graphs does not satisfy non-holonomic constraints, state lattice method has been proposed [3]. Paper [4] investigates using a lattice of *spatiotemporal* states which uniformly cover the surface of the road being traversed, and employ smooth cubic curvature polynomials, for finding feasible continuous plans. Advancing this method to use quartic curvature polynomials and trajectory optimization with respect to a linear summation of some cost functions has been studied in [5].

Complete planning methods face the challenge of choosing a suitable resolution value for discretization stage. Higher resolutions require tremendous amount of memory and heavy computations which make it inappropriate for real-time decision making. On the other side, reducing it, affects the quality of output.

In contrast, random sampling methods are suitable for planning in higher dimension domains, as they do not require the explicit enumeration of planning constraints. RRT is the most well-known method in this class [6].

Forward simulation in a system with complex nonlinear dynamics has been discussed in [7] by the help of closed-loop control, where a planner based on RRT for an autonomous vehicle is designed and implemented for the first time (CL-RRT). *Online replanning* mechanism, introduced in this method, reuses the controller inputs stored in the tree and performs a *repropagation* from the latest vehicle state. This mechanism takes a lot of computation for each cycle of decision making procedure.

A typical deficiency of above methods is failure in finding an optimal plan; a problem which is resolved by using RRT* method [8]. The output of this planner converges asymptotically to the optimal plan. This algorithm also has been customized for dynamic systems with non-holonomic constraints in [9].

In general, the major drawback of random sampling method is low quality of output plans. Ignoring information related to shape and size of obstacles in the environment, results in non-smooth trajectories which include unnecessary motions. This situation becomes more noticeable in narrow passages.

In [10] spline curves are employed for producing smooth paths with continuous curvature. This path planner is suitable for non-holonomic robots.

Additionally sampling from the space of trajectories, instead of configuration space is studied in [11]. Using cross-entropy method, in one step, some trajectories are generated from a distribution and their cost is computed. And in another approach, the distribution is updated from a high-quality subset of these trajectories. Then recursively, the two phases are repeated until an optimum plan is achieved.

Additionally, there are reactive motion planners, which propose a motion to the robot without producing any motion plan for achieving the destination. Ability to react quickly to changes in the environment is an advantage of such methods. Even though many of planners in the past have been using these algorithms because of their simplicity, those idea can be employed in new autonomous systems to reduce the time of motion planning significantly.

In [12] potential field has been used to speed up the procedure of making tree in RRT* algorithm. In [13] the forces in potential fields are redefined by fuzzy TSK and Mamdani models. Fast marching method in [14] inspired by the pattern of waves spread on water surface, generally liquids, caused by throwing an object in the water, is capable of finding a path in a linear time with respect to the configuration space size. This method overcomes the problem of local minimum of potential fields. However, it suffers from problems engendered by discretization.

Searching for an optimal plan is also another important challenge. Generating optimal plans through random sampling has been proposed in [15]. Additionally, evolutionary approaches have demonstrated high efficiency in solving non-convex and non-classical optimization problems. The Paper [16] reviews the applications of these methods in motion planning.

In this paper we present a method based on Artificial Potential Fields for motion planning problem. Our planner reacts immediately to sudden changes like obstacle movements. Besides, the proposed method is capable of improving the quality of output plans by automatically modification of its parameters in the environment. The paper is structured as follows:

In section 2 a short background about artificial potential field is presented. The proposed method will be described in sections 3 and 4, where they describe the idea of using the information of an available path, in motion planning with artificial potential fields, and a mechanism for improving the quality of output plans, respectively. Section 5 is dedicated to presentation of experimental and simulation results. Finally in section 6, we discuss the conclusion.

II. MOTION PLANNING WITH ARTIFICIAL POTENTIAL FIELDS

In an artificial potential field each position has a potential value which determines the level of energy for that point. When a charged particle, or in other words a planning agent, is put in the field, it moves in direction of decreasing the potential field; which is the reverse direction of the gradient of potential function.

$$F(s) = -\nabla U(s) = -\sum_i \nabla U_i(s) \qquad (1)$$

Hence the applied force equals the summation of forces in the field caused by different sources e.g. obstacles and the destination.

A motion plan can be produced from a potential field by starting from the initial state, calculating the force vector in each state and finding the next state of plan using a transfer function, until it reaches to goal state or get trapped in a local minimum. Transfer function is determined based on the motion behavior of the planning agent and dynamic and non-holonomic constraints are important in defining this function. We also need to discretize the planning process in time dimension and determine the time step between the stations of the plan.

In artificial potential field planning algorithm, that we call it Naive APF, force sources include attractive and repulsive forces. The attractive force is applied in the direction of goal point to an object and its magnitude is proportional to distance between the object and goal. The repulsive force applies to an object from any present obstacle in the environment and tends to keep the robot away from the obstacle. The intensity of the force is inversely proportional to the quartic distance between the robot and an obstacle. Then the formula of applied force in naive APF is as follows:

$$\begin{aligned}F^{tot}(s) &= \alpha\, F^{att}(s) + \sum_{i=1}^{m} \beta_i\, F^{rep}_{ob_i}(s) \qquad (2)\\ &= \alpha\,(s_G - s) \\ &\quad + \sum_{i=1}^{m} \beta_i \frac{s - s_{ob_i}}{\|s - s_{ob_i}\|^4}\end{aligned}$$

where $\alpha$ and $\beta_i$ s are coefficients for forces, $s_G$ is the goal state, and $s_{ob_i}$ is the nearest point in obstacle i to state $s$. The major deficiency of this method, i.e. Naïve APF, is getting trapped in local minima of potential function and failure in producing a plan, even when an actual path exists. The main cause of this phenomenon is greedy behavior of approaching to goal point that we will address in the next section.

As it is shown in the above equation, the output plan will be significantly influenced by changing one or some coefficients of the forces, i.e. $\alpha$ and $\beta_i$s that we will address them with force weights. As depicted in Fig. 1 by increasing the coefficient of repulsive force, motion plans will be produced each with a different amount of length, smoothness and safety. In section 4, it will be shown that a better plan can be achieved by finding the optimal weights.

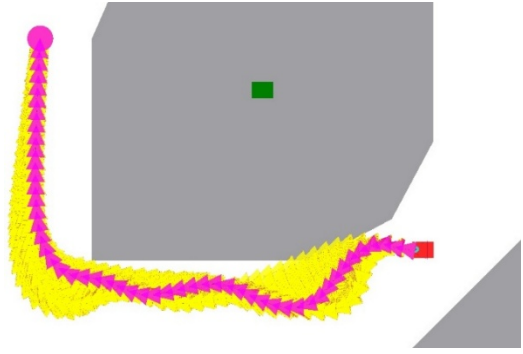

Fig. 1. Effect of the changes in magnitude of repulsive force on result of APF output. Highlighted red plan shows the best plan in this spectrum.

### III. UTILIZING A PRIOR PATH IN APF METHOD

The attractive force alone does not satisfy the requirements for navigation of robot toward any goal. To reach the main goal the robot should traverse a sequence of sub-goals in its way. This is the key idea for defining a new source of force that directs robot to the goal region through some sub-goals. Thus we have defined a new type of force and named it directive force.

First assume that we have a collision-free path between the initial and goal state, namely prior path, including a sequence of segments, which does not necessarily satisfy the constraints of motion planning problem. We first declare a distance criterion to find out the distance of a point, $s$, from each of the segments, $l_i$, on the prior path. The directive force then will be applied in the direction of the nearest segment. The distance function that we use is as follows:

$$d(s, l_i) = \frac{\|s - v_i^s\| + \|s - v_i^e\|}{\|v_i^e - v_i^s\|} \quad (3)$$

Where $v_i^s$ and $v_i^e$ denotes the start and end nodes of the $i$'th segment i.e. $l_i$. Using this distance function, the locus of points which are equidistant from a segment forms an elliptic region. Then bisection of two sequent segments determines the boundary of regions, each of which guides the forces in one direction. The directive force is just directed by the nearest segment and is a unit vector parallel to that segment according to the following formula:

$$F^{dir}(s) = \sum_i \Gamma(s, l_i) \frac{v_i^e - v_i^s}{\|v_i^e - v_i^s\|} \quad (4)$$

Where $\Gamma$ is a function that eliminates the effect of all segments except the nearest one:

$$\Gamma(s, l_i) = \begin{cases} 1 & \operatorname*{argmin}_i d(s, l_i) \\ 0 & o.w. \end{cases} \quad (5)$$

Fig. 2 illustrates how an environment is partitioned by bisections of line segments.

Now assume that we only use the directive force in a potential field, starting to produce a motion plan without dynamic and non-holonomic constraints, the final plan will be just fitted on the prior path.

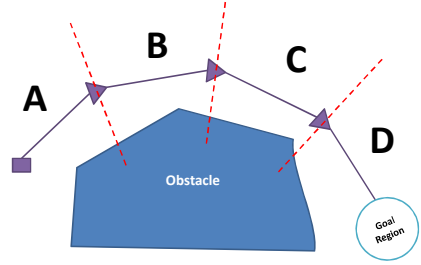

Fig. 2. Dividing the configuration space into 4 parts by 4 segments of a prior path

However there are two reasons behind defining this force. Firstly we intend to take the constraints of the planning agent into account. Then we can produce a feasible plan for our robot from a prior path which may be infeasible to traverse. Another interesting result will be achieved by combining repulsive and directive forces together. In other words the repulsive forces increase the distance of the plan from obstacles that leads to a safer plan.

We should add this point that we do not confine the algorithm to use any specific path planning algorithm and every path planning approach which generates a sequence of sub-targets or path segments can be employed, however we chosen RRT method because of its fast speed in finding a path.

Moreover, we present a technique to limit the search space, in order to increase the speed of path finding and density of tree nodes around previous paths that provide possibility of generating high-quality paths in the next search iterations.

Suppose that the sampling space is a closed region like $C_k$ when a plan with length $l_k$ is obtained in the $k$'th iteration of the planning algorithm, a limitation may be added to the region and therefore make it smaller, if it is possible. Since an upper bound for the sum of Euclidean distance of each point on the path from the origin and goal point is $l_k$, there is no need to search distances any further than $l_k$ from these two points. Hence an elliptic shape area having origin and goal points $F_1$ and $F_2$ as its focal points and $l_k$ as its bigger diameter specifies the search region. In the next iteration of planning we will consider the intersection region of $C_k$ and $E_k$, namely $C_{k+1}$ for randomized sampling. The process is illustrated in Fig. 3.

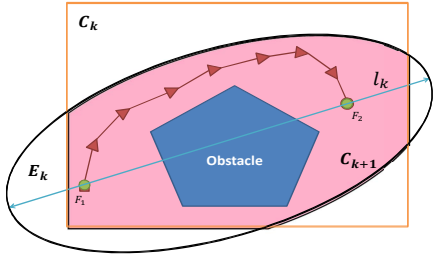

Fig. 3. Intersecting a closed region with the elliptic region to construct next iteration's sampling area. The pink area shows the search space for step $k+1$

## IV. OPTIMIZATION OF MOTION PLANS

There are several parameters in our method which significantly influence the quality of the output plan, therefore we decided to use an optimization mechanism to find lower cost motion plans, in other words we are willing to explore the space of motion plans in order to obtain more fitted solutions. This process will be performing while the robot is traversing its path toward the destination and does not require the robot to stop until finishing optimization process. In the proposed algorithm the planner will produce new plans in each cycle and choose the best one for running in any time.

A complete set of cost functions for motion planning is presented in [17] while we will just consider the three most important ones: length, safety and smoothness. In our approach an evolutionary algorithm is employed to modify the available plans and produce new solutions. As it is illustrated in Fig. 4 a unit is responsible for generating prior paths between the origin and destination. The outputs of this step fills a bank of paths which thereof are employed as prior paths for the main motion planning algorithm. The choice of a prior path is based on roulette wheel mechanism where the probabilities assignment is reversely related with the path length. We also eliminate the older paths when they become invalid according to new changes in the configuration space.

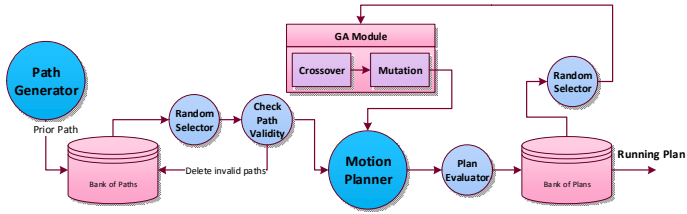

Fig. 4. The structure of optimization process for producing optimal plan

As we described the main cycle of the proposed motion planner, it needs a prior path to produce directive forces. Hence we initialize the bank of prior paths with the most trivial prior path i.e. a path including just one segment connecting the initial state to the goal state. Thus our algorithm will act like Naïve APF until no prior path is found by path generator module. Moreover we use a multi-objective evaluator to assess the produced plans. In this evaluator the summation of three objective functions will be calculated for each plan. Also in the individual fitness, the crowding distance is considered to diversify the produced plans. This obtained diversity uplifts the algorithm's convergence rate and additionally, helps to find a feasible plan in the bank, when some earlier plans cease to be feasible because of changes in the configuration space. In the evolution unit, uniform crossover and Gaussian mutation are employed to generate new offspring. A chromosome in this genetic problem is a vector of the forces weights. In other words it contains all repulsive force coefficients and also the directive force coefficient. We are also using the elitism operator to guarantee the optimization convergence.

## V. SIMULATION AND EXPERIMENTAL RESULTS

We discuss our implementations and experiments in this section. To test our planner, we employed a simulated autonomous vehicle in Webots software. Under the assumption that a complete map of the environment is available to decision system, the structure of streets is modeled by various obstacles that surround the free space and are fed to the planner. Only cars and small objects which may exist on the streets will be detected by the range finder sensor installed on the robot and the planner will regard such detected obstacles. The planning environment is a rectangle of size 300*300 meter out of which 77.5% is covered by obstacles and the rest forms the streets.

The simulated vehicle itself covers an area of 2.5*5 $m^2$. The planner is developed in C++. For collision checking, in planning stage, Box2D library is employed [18]. The planner was run on a 2.4 GHz core 2 Duo and 4 GB of memory. Fig. 5 shows a screenshot of the simulation environment in Webots.

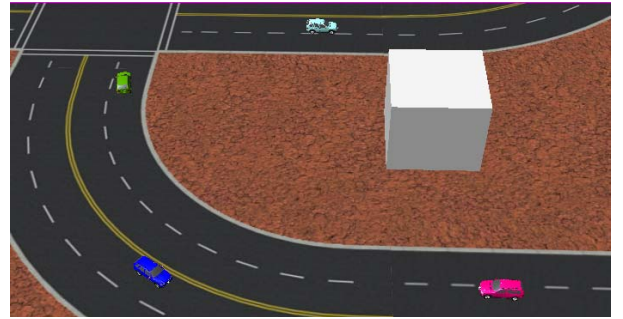

Fig. 5. The simulation environment for autonomous vehicle in Webots PRO.

As mentioned before we use RRT algorithm to generate RRT paths. Fig. 6 illustrates the RRT tree and the obtained path. It shows how the number of spurious nodes in the tree decreased by applying our method to confine the search area.

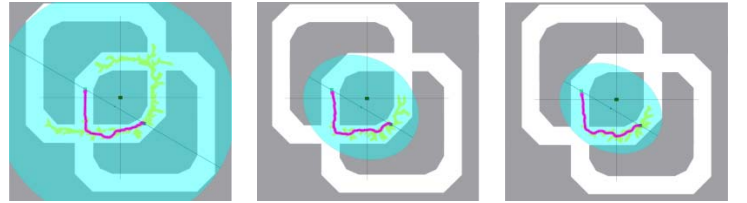

Fig. 6. The result of limiting the sampling area by an elliptic shape. From left to the right the search area diminishes thus the random tree will contain fewer nodes.

We designed a planning scenario to test our assumptions, running the path generator and confining the search area in each iteration. This scenario was repeated for 50 times to observe the number of nodes in the random tree in each iteration of each run to find the average behavior of the algorithm. We observed that a major decrease in the number of tree nodes happens in the second iteration and then the elliptic area does not diminish significantly. We depicted the results in Fig. 7.

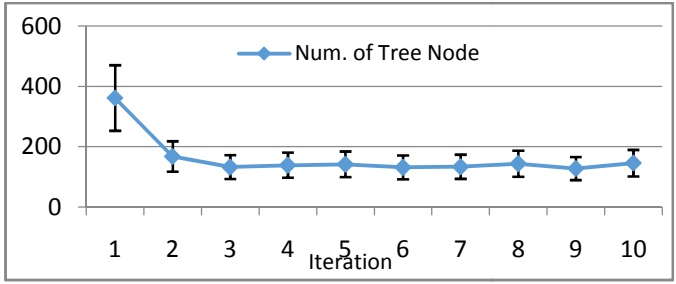

Fig. 7. Average number of the nodes in random tree in 10 iterations, by limiting the sampling area

In Fig. 8. Three(a) and Fig. 8(b) the attractive and repulsive forces are depicted respectively. Incomplete output of Naïve APF approach which is obtained by calculating the sum of these forces in each state, is depicted in Fig. 8(c). This motion plan is trapped in a local minimum where any further steps will cause a collision with one of the obstacles. In Fig. 8(d) the primary result of RRT algorithm is depicted. This path is simplified to the one in Fig. 8(e). You can see that how directive forces are distributed in the configuration space. Finally the result of our method is shown in Fig. 8(f). This plan is completely smooth and feasible for robot.

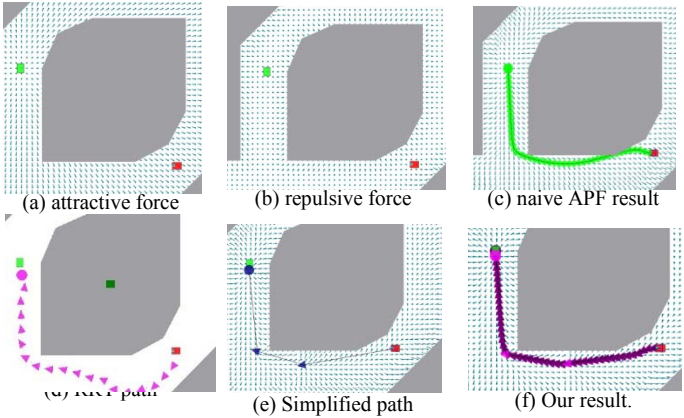

(a) attractive force   (b) repulsive force   (c) naïve APF result

(d) RRT path   (e) Simplified path   (f) Our result.

Fig. 8. Three different types of forces in the potential field.

Fig. 9 demonstrates how the entire motion planning process time is raised with respect to path planning time, i.e. the time which is needed to generate a prior path with RRT method. In this experiment, various scenarios with different distances are designed to test the running time of the proposed algorithm. An average of 4 to 14 milliseconds increase in this time is because of plan production through potential field.

One of the prominent features of the proposed method is the ability of fast reaction when the robot encounters sudden changes in the environment. While most of the available motion planning methods take many times to repair the current plan the reactive property of proposed method leads in a quick response. For testing this hypothesis, we tested our method versus the Naive RRT algorithms. In this experiment, we added a new obstacle to the environment and changes its position, just upon the current plan and waited for reconstructing a new valid plan by the algorithm. Fig. 10 shows the results of in which this task has been repeated for different situations.

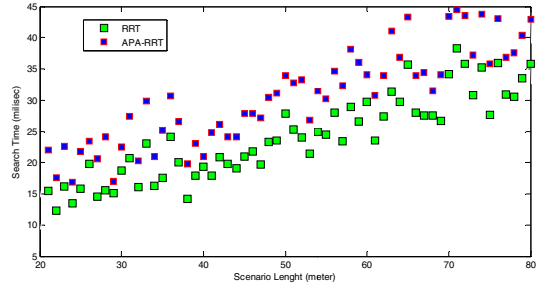

Fig. 9. Comparing the total time of running proposed algorithm (red) with RRT algorithm (green) in scenarios with different lenghts.

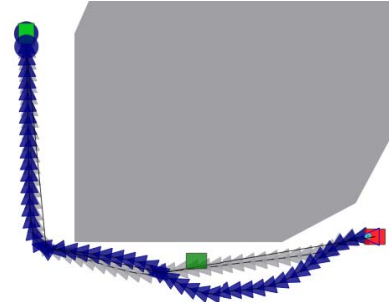

Fig. 10. The result of putting a new obstacle on current plan of the robot. The planner quickly changes the field forces to generate the new plan.

In the above experiment, the average time for running the method for 20 tests is 7.24 millisecond. But the average time for running RRT algorithm again takes about 17.15 millisecond.

Another major part in our work is proposing a new mechanism for evolving the plans towards optimum plan as described in the previous section. We have considered a bank of prior paths with a capacity of 100 paths, such generated paths will put in this bank, after evaluating their lengths. These paths are sorted in increasing order. The population size of the genetic algorithm is set to 200 individual and crossover probability and mutation probability were set to 0.75 and 0.15, respectively.

In Fig. 11(a) the convergence of the algorithm is depicted and in Fig. 11(b) it is shown that how our proposed method surpass other planning methods in producing lower cost solutions. The cost of the output plan of Naïve APF remains constant because it is a deterministic approach. RRT algorithm on the other hand is kept running for a while and we see that better plans were found by repeating the algorithm. We see that our method provides a better performance in finding better solutions in a short time.

## VI. Conclusion

This paper presented the APA-APF planning method, an extension of standard APF planner, which uses prior paths between origin and destination to free the potential field from local minima. Adaptive feature of this planner has been established through an evolutionary multi-objective optimization mechanism. Even though the method is customized for autonomous driving, however its simplicity makes it suitable for other domains. In another word, instead of using traditional sampling-based planning algorithms or systematic searches in connectivity graphs, our approach meets the needs of real-time motion planning, particularly in mobile robots. We also showed how the idea of limiting the sampling area in path re-planning stage, speed up finding new solutions by reducing the number of generated nodes in the random tree. The method is implemented in C++ and is tested on a simulated autonomous vehicle in Webots simulation environment.

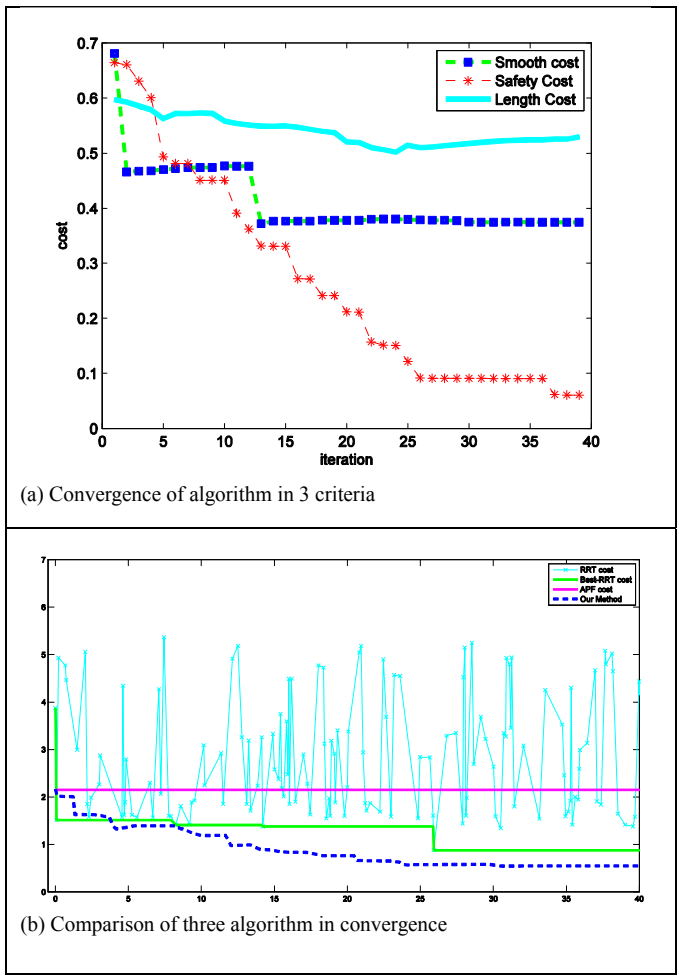

(a) Convergence of algorithm in 3 criteria

(b) Comparison of three algorithm in convergence

Fig. 11. Results of running proposed algorithm.